\documentclass[pdflatex,sn-mathphys-num]{sn-jnl}

\usepackage{graphicx}%
\usepackage{multirow}%
\usepackage{amsmath,amssymb,amsfonts}%
\usepackage{amsthm}%
\usepackage{mathrsfs}%
\usepackage[title]{appendix}%
\usepackage{xcolor}%
\usepackage{textcomp}%
\usepackage{manyfoot}%
\usepackage{booktabs}%
\usepackage{bm}
\usepackage{algorithm}%
\usepackage{algorithmicx}%
\usepackage{algpseudocode}%
\usepackage{listings}%

\usepackage{xspace}
\newcommand{\ivttriplet}{\texttt{$\langle$instrument, verb, target$\rangle$}\xspace}

\newcommand{\triplet}[3]{\texttt{$\langle$#1, #2, #3$\rangle$}\xspace}

\newcommand{\anatprior}{weak anatomy\xspace}
\newcommand{\Anatprior}{Weak anatomy\xspace}

\newcommand{\mAP}{\textrm{mAP}}

\newcommand{\figref}[1]{\figurename~\ref{#1}}
\newcommand{\tabref}[1]{\tablename~\ref{#1}}
\newcommand{\secref}[1]{Section~\ref{#1}}

\newcommand{\mysubsection}[1]{\subparagraph{#1}}

\theoremstyle{thmstyleone}%
\theoremstyle{thmstyletwo}%

\theoremstyle{thmstylethree}%

\begin{document}
\title[Article Title]{Grounding Surgical Action Triplets with Instrument Instance Segmentation: A Dataset and Target-Aware Fusion Approach}


\author[1]{\fnm{Oluwatosin} \sur{Alabi}}\email{oluwatosin.alabi@kcl.ac.uk}

\author[1]{\fnm{Meng} \sur{Wei}}\email{	meng.wei@kcl.ac.uk}

\author[1]{\fnm{Charlie} \sur{Budd}}\email{charles.budd@kcl.ac.uk}

\author[1]{\fnm{Tom} \sur{Vercauteren}}\email{	tom.vercauteren@kcl.ac.uk}

\author*[2]{\fnm{Miaojing} \sur{Shi}}\email{mshi@tongji.edu.cn}

\affil[1]{
\orgdiv{BMEIS School}, 
\orgname{King's College London}, 
\city{London},  \country{United Kingdom}}

\affil[2]{
\orgdiv{EIE College},
\orgname{Tongji University}, 
\city{Shanghai}, \country{China}}



\abstract{\textbf{Purpose:} Understanding surgical instrument–tissue interactions requires not only identifying which instrument performs which action on which anatomical target, but also grounding these interactions spatially within the surgical scene. Existing surgical action triplet recognition methods are limited to learning from frame-level classification, failing to reliably link actions to specific instrument instances. Previous attempts at spatial grounding have primarily relied on class activation maps, which lack the precision and robustness required for detailed instrument–tissue interaction analysis.
To address this gap, we propose grounding surgical action triplets with instrument instance segmentation, or \emph{triplet segmentation} for short, a new unified task which 
produces spatially grounded \ivttriplet outputs.

\textbf{Methods:} We start by presenting CholecTriplet-Seg, a large-scale dataset containing over 30,000 annotated frames, linking instrument instance masks with action verb and anatomical target annotations, and establishing the first benchmark for strongly supervised, instance-level triplet grounding and evaluation. 
To learn triplet segmentation, we propose TargetFusionNet, a novel architecture that extends Mask2Former with a target-aware fusion mechanism to address the challenge of accurate anatomical target prediction by fusing \anatprior priors with instrument instance queries. 

\textbf{Results:} Evaluated across recognition, detection, and triplet segmentation metrics, TargetFusionNet consistently improves performance over existing baselines, demonstrating that strong instance supervision combined with weak target priors significantly enhances the accuracy and robustness of surgical action understanding.

\textbf{Conclusion:} Triplet segmentation establishes a unified framework for spatially grounding surgical action triplets. The proposed CholecTriplet-Seg benchmark and TargetFusionNet architecture pave the way for more interpretable, fine-grained surgical scene understanding.
}


\maketitle

\section{Introduction}\label{sec::introduction}

\begin{figure}[b]
    \centering
    \includegraphics[width=0.95\linewidth]{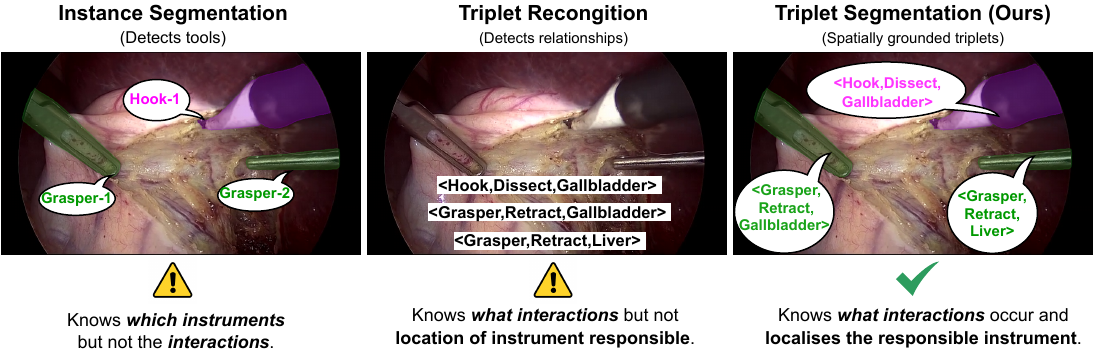}
    \caption{
        Comparison of instance segmentation, triplet recognition, and triplet segmentation. Triplet segmentation grounds \ivttriplet triplets in space, unifying triplet recognition and instrument localisation for interpretable surgical scene understanding. 
    }
    \label{fig:instance_recognition_triplet_segmentation}
\end{figure}

Modelling surgical instrument–tissue interactions is essential for understanding surgical actions and workflows. Action triplets \ivttriplet have emerged as a concise yet powerful representation for instrument-tissue interaction, capturing interactions such as \triplet{Grasper}{Grasp}{Gallbladder} or \triplet{Scissors}{Cut}{Cystic\_duct} \cite{nwoye2020recognition}. However, existing triplet recognition methods \cite{nwoye2020recognition,nwoye2022rendezvous,yamlahi2023self} operate at the frame level, predicting which triplets are present in a scene without identifying the location as seen in \figref{fig:instance_recognition_triplet_segmentation}. This lack of spatial grounding makes predictions ambiguous and reduces their clinical value, as precise action localisation is crucial for surgical skill assessment, context-aware assistance, and workflow analysis.

In this work, we propose grounding surgical action triplets with instrument instance segmentation, a task concisely referred to as \emph{triplet segmentation} that unifies instrument instance segmentation with verb and target classification to produce instance-grounded \ivttriplet outputs (see Fig.~\ref{fig:instance_recognition_triplet_segmentation}). 
Unlike frame-level recognition or coarse bounding-box detection,  triplet segmentation explicitly links each predicted triplet to a corresponding pixel-level instrument instance mask. This provides fine-grained interpretable insights into surgical activities, revealing both actions and their precise spatial context.

However, existing datasets are not designed for this task: they either lack instance-level masks, do not provide complete triplet labels, or are not spatially aligned. 
To address this gap, we present CholecTriplet-Seg, the first large-scale dataset to pair high-quality instrument instance masks with action verb and anatomical target annotations. 
Building on CholecInstanceSeg \cite{alabi2025cholecinstanceseg} and the CholecT50 triplet labels \cite{nwoye2022rendezvous}, CholecTriplet-Seg provides over 30,000 annotated frames from 50 laparoscopic videos, linking each instrument instance to its associated action and anatomical target. 
This dataset establishes a benchmark for triplet segmentation and enables fine-grained, instance-level interaction analysis, forming a solid foundation for future extensions beyond cholecystectomy. 
To measure performance on this dataset, we introduce Triplet Segmentation mAP, a metric that extends existing triplet evaluation protocols to the instance level. It assesses both triplet-level and component-level accuracy, requiring models to correctly classify the triplet and accurately localise the corresponding instrument instance.

Achieving reliable anatomical target prediction remains a major bottleneck. Anatomy is inherently ambiguous and difficult to annotate at the pixel-level, even for expert surgeons. 
Manual anatomy segmentation masks are also prohibitively expensive to produce. 
Errors in target prediction usually drive overall triplet accuracy because a triplet is only correct if all three components in \ivttriplet are correct. To address this challenge, we propose TargetFusionNet, a novel architecture that extends Mask2Former \cite{cheng2022masked} with a target-aware fusion mechanism. TargetFusionNet integrates \anatprior logits obtained from a pretrained network into the transformer decoder, improving the association between instruments and their likely targets. We demonstrate that this approach significantly improves target recognition and overall triplet accuracy compared to the existing CAM-based baseline, a standard instance segmentation baseline, and early / late fusion strategies.

Our key contributions are as follows:
\begin{itemize}
\item Task: We define triplet segmentation, which unifies instrument instance segmentation and action–target classification for spatially grounded triplet analysis.
\item Dataset: We release CholecTriplet-Seg, the first large-scale dataset linking instrument instances with verbs and targets for fine-grained triplet grounding.
\item Metric: We introduce Triplet Segmentation mAP, a metric that evaluates both component and triplet-level accuracy with instance-level localisation.
\item Method: We propose TargetFusionNet, a target-aware segmentation architecture that  addresses target prediction bottlenecks, improving triplet grounding and establishing a new benchmark on our dataset.
\end{itemize}

\section{Related work}\label{sec::related_works}

\mysubsection{Surgical triplet recognition (classification)} 
Most prior works treat triplet recognition as a frame-level multi-label classification task, focusing on identifying which triplets are present rather than which instrument executes the action. 
TripNet \cite{nwoye2020recognition} and Rendezvous (RDV) \cite{nwoye2022rendezvous} introduced attention-based models guided by weak spatial priors from class activation maps (CAMs), achieving strong frame-level recognition but lacking spatial grounding. 
Later studies explored disentangled classification \cite{chen2023surgical}, self-distillation \cite{yamlahi2023self}, and temporal reasoning \cite{sharma2023rendezvous}, yet none localise which instrument performs each action. 
Existing datasets such as CholecT50 \cite{nwoye2022rendezvous} remain limited to frame-level triplet labels without explicit spatial alignment.

\mysubsection{Surgical triplet detection (bounding box grounding)}
Triplet detection extends recognition by introducing bounding box spatial grounding. 
The CholecTriplet2022 Challenge~\cite{nwoye2023cholectriplet2022} formalised this via bounding-box localisation from weak supervision. 
Subsequent work focused on refining frame-level recognition through disentangled classification \cite{chen2023surgical}, self-distillation \cite{yamlahi2023self}, and temporal reasoning \cite{sharma2023rendezvous,pei2025instrument} , but spatial grounding remains coarse since triplets are not explicitly linked to instrument instances. 
Bounding boxes also only offer poor localisation for elongated instruments in non axis-ligned orientation \cite{han2025robust}.

\begin{table*}[t]
\centering
\caption{
Overview of the datasets involved in constructing CholecTriplet-Seg. 
CholecT50 provides frame-level \ivttriplet labels, 
CholecInstanceSeg provides dense instrument masks, and our CholecTriplet-Seg dataset unifies both by matching triplet annotations with instrument instances.
}
\begin{tabular}{l|c|c|c}
\hline
Dataset & Videos & Annotation Type & \# Frames \\
\hline
CholecT50~\cite{nwoye2022rendezvous} & 50 & Frame-level triplet labels  & 100,861 \\
CholecInstanceSeg~\cite{alabi2025cholecinstanceseg} & 85 & Instrument instance masks  & 41,933 \\
CholecTriplet-Seg (Ours) & 50 & Triplet + instance (aligned) & 30,955 \\
\hline
\end{tabular}
\label{tab:dataset_overview}
\end{table*}

\mysubsection{Instrument instance segmentation datasets}
Research on surgical instrument segmentation has progressed from early binary or semantic labelling towards large-scale datasets offering fine-grained, instance-level annotations \cite{ahmed2024deep}. Recent efforts such as PhaKIR \cite{rueckert2025comparative}, GraSP \cite{ayobi2024pixelwise}, and CholecInstanceSeg \cite{alabi2025cholecinstanceseg} have established comprehensive resources for training and benchmarking instrument instance segmentation models. 
Among these, CholecInstanceSeg provides high-quality instrument masks aligned with CholecT50 \cite{nwoye2022rendezvous}, forming the foundation for our CholecTriplet-Seg dataset that unifies instance segmentation with verb–target labels for grounded triplet analysis.

\section{Material: The CholecTriplet-Seg dataset}\label{sec::cholectriplet_seg_dataset}
CholecTriplet-Seg is a dataset of laparoscopic cholecystectomy videos with instrument-instance-grounded \ivttriplet annotations. 
It is created by combining frame-level triplet labels from CholecT50 \cite{nwoye2022rendezvous} with dense instrument masks from CholecInstanceSeg \cite{alabi2025cholecinstanceseg}, synchronising corresponding frames and linking each instrument instance to its verb and target labels (see~\tabref{tab:dataset_overview}). 

\mysubsection{Triplet schema}
CholecT50 defines an annotation schema of six instruments, ten verbs, and fifteen targets, with 100 clinically valid triplet combinations established through expert consensus; we directly adopt this schema for CholecTriplet-Seg.

\mysubsection{Alignment and quality control and ethics approval}
Frames from both datasets are aligned using timestamps to associate each instrument mask with a corresponding \ivttriplet label. 
Automatic alignment was refined through manual verification to correct missing or ambiguous mappings, add absent masks and triplets, and ensure consistent assignments. 
Two rounds of visual inspection confirmed annotation quality, with additional statistics and details provided in the appendix. 
The final dataset contains 30{,}955 annotated frames and 49{,}866 spatially grounded triplets from 50 cholecystectomy videos, including 15 fully annotated and 35 sparsely sampled cases (1 in 30 frames). 
Training, validation, and test splits follow the CholecInstanceSeg protocol. 
No ethics approval was required as only existing public datasets were used with additional annotations.

\section{Methods}
Our goal is to predict spatially grounded \ivttriplet outputs by linking each instrument instance to its corresponding verb and target labels. We formalise this as the triplet segmentation task: given an input image $x \in R^{H \times W \times 3}$, the model outputs a set of instance masks $M = \{ M_k \}$, each associated with a triplet label:
\[
    \mathcal{T} = \{ \langle I_k, v_k, t_k \rangle \mid M_k \in M \},
\]
where $I_k$, $v_k$, and $t_k$ denote the instrument, verb, and target class of instance $k$.

While we have accurate spatial annotations for instruments, pixel-level annotations for anatomical targets are unavailable. This makes target prediction a major bottleneck in triplet segmentation. Our evaluation on CholecTriplet-Seg (see \emph{Results}) shows that even with correct instrument masks and verb labels, poor target classification substantially limits Triplet $\mAP$ (see \emph{Evaluation and Metrics}). Anatomical targets are particularly challenging.
They often exhibit subtle appearance differences, lack distinct boundaries, and are inconsistently defined, even among expert surgeons \cite{owen2022automated}.
To address this, we propose TargetFusionNet, a model that improves target classification by incorporating \anatprior priors derived from a pre-trained tissue segmentation network. These predictions are not treated as ground truth but as weak spatial cues that help disambiguate instrument–target interactions.

Targets represent anatomical structures (e.g., liver, gallbladder, cystic duct) that lack the clear boundaries and distinctive features of surgical instruments. Pixel-level target annotations for our triplet categories are unavailable because producing them would require extensive expert labelling and consensus-building. Without explicit target masks, target classification remains a major source of error.

\mysubsection{\Anatprior priors}
We incorporate coarse anatomical cues using inference logits from EndoViT~\cite{batic2024endovit}, trained on CholecSeg8k \cite{hong2020cholecseg8k} for tissue segmentation in cholecystectomy surgeries. EndoViT predicts six broad tissue classes that only partially overlap with the 15 anatomical target categories in our triplets, introducing class mismatches and noisy predictions. Nevertheless, these logits capture useful spatial context, such as rough anatomical boundaries (e.g., distinguishing liver regions from the gallbladder). We treat them as noisy priors rather than ground truth and integrate them into TargetFusionNet to guide target prediction.
To avoid data leakage, we follow the CholecInstanceSeg split protocol \cite{alabi2025cholecinstanceseg}, ensuring that any CholecSeg8k frames overlapping with CholecT50 appear only in the training partition. This prevents shared visual content between training and test sets, guaranteeing fair evaluation of anatomical priors.

\begin{figure}[t]
    \centering
    \includegraphics[width=\linewidth]{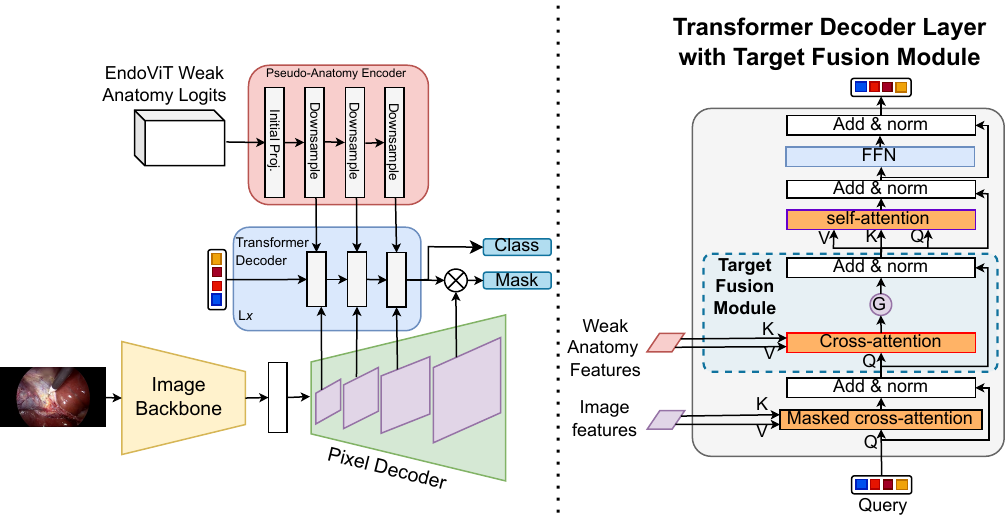}
    \caption{
        Overview of the TargetFusionNet architecture (left) and the Transformer Decoder Layer with Target Fusion Module (right). 
        TargetFusionNet augments Mask2Former with an additional encoder that processes \anatprior logits from EndoViT into multi-scale feature maps. 
        These \anatprior features serve as auxiliary memory for the transformer decoder. 
        Each transformer decoder layer contains a Target Fusion Module, where instance queries interact with \anatprior features via gated cross-attention. 
        This allows the model to selectively incorporate coarse anatomical priors while refining triplet predictions.
        }
    \label{fig:target_fusion}
\end{figure}

\mysubsection{TargetFusionNet}\label{sec:targetfusionnet}
Our TargetFusionNet, illustrated in \figref{fig:target_fusion}, builds on Mask2Former \cite{cheng2022masked}, a transformer based instance segmentation model. 
We enhance its architecture with a target-aware fusion mechanism with three components:
\begin{itemize}
    \item \Anatprior encoder: The EndoViT logits are projected to the same feature dimension as the transformer decoder queries. A series of downsampling convolutions converts them into multi-scale \anatprior feature maps that act as anatomical priors.
    
    \item Transformer decoder layers with target fusion: Mask2former transformer decoder layers are augmented with a target fusion module which attends to the \anatprior features via a gated cross-attention as in \figref{fig:target_fusion}. This allows the decoder to selectively incorporate anatomical context while refining triplet representations.
    
    \item Gated cross-attention: Each transformer query \( Q \) attends to \anatprior features \( (K_t, V_t) \) through a learnable gating mechanism:
    \[
    Q' = Q + \sigma(W_g A) \odot A,
    ~ \mathrm{where} ~  A = \mathrm{Attn}(Q, K_t, V_t),
    \]
    where \( \sigma \) is a sigmoid function, \( W_g \) is a learnable linear layer, and \( A \) is the output of cross-attention between \( Q \) and the \anatprior features. The gate controls how much anatomical context is injected, preventing noisy priors from dominating visual features.

\end{itemize}
%

\mysubsection{Triplet output and loss}
Following \citet{nwoye2022rendezvous}, we represent each \ivttriplet combination as a single class within a 100-way classification space defined by the clinically valid triplets of CholecT50.
For each predicted instance mask $M_k$, TargetFusionNet directly outputs a single triplet label $c_k \in {[1; \dots; 100]}$, where each class $c_k$ uniquely corresponds to a specific triplet combination. 
Each transformer query predicts one triplet label, while mask generation still follows the standard Mask2Former query–pixel interaction. The target-fusion gate acts only on the query embeddings before classification, enriching triplet semantics without altering the mask prediction process.
The network is trained end-to-end using standard Mask2Former losses. A comparison  of models with multitask loss variants, where separate losses are applied to instrument, verb, and target predictions,  is included in the appendix.

\section{Experiments}\label{sec:experiments}
We evaluate TargetFusionNet on the proposed CholecTriplet-Seg benchmark and compare it against both weakly and strongly supervised baselines to quantify the benefits of strong instance supervision and target-aware fusion.  

\mysubsection{Baselines}
The following models are used for comparison:
\begin{itemize}
    \item RDV-Det~\cite{nwoye2023cholectriplet2022}: the official CholecTriplet2022 baseline extending the RDV framework with class activation maps (CAMs) for weak localisation. It predicts triplets at the frame level and assigns them to instruments via CAM-derived bounding boxes, offering only coarse spatial grounding.  
    \item RDV + Mask2Former: augments RDV-Det by replacing CAMs with instance masks predicted by a Mask2Former model trained on CholecTriplet-Seg. This isolates the benefit of strong instance segmentation while retaining RDV’s triplet association mechanism.  
    \item Mask2Former-Triplet: extends Mask2Former with a 100-class triplet head that jointly predicts instance masks and triplet labels, removing heuristic matching and providing a strong baseline for comparison with TargetFusionNet.  
\end{itemize}

\mysubsection{Evaluation protocol}
Performance is assessed across triplet recognition (classification), triplet detection (bounding box grounding), and triplet segmentation settings to capture both triplet correctness and spatial grounding. Following the CholecT50 protocol~\cite{nwoye2023datasplitsmetricsmethod}, we report the standard recognition and detection mAP (bounding box based). 
To enable instance segmentation level evaluation, we introduce a new \emph{triplet segmentation mAP} metric, which extends existing triplet evaluation to the pixel level by requiring both correct triplet classification and accurate mask localisation. This metric provides a unified framework for comparing models across frame, box, and mask-level grounding, thereby establishing the first benchmark for spatially grounded triplet segmentation.
Metrics are reported for individual components ($\mAP_I$, $\mAP_V$, $\mAP_T$), component pairs ($\mAP_{IV}$, $\mAP_{IT}$), and full triplets ($\mAP_{IVT}$), with superscripts $rec$, $det$, or $seg$ indicating the evaluation mode (e.g., $\mAP_{IVT}^{seg}$). Further details on metric computation are available in the appendix and accompanying codebase.  
We also present qualitative results, visualising predicted masks and triplets to highlight common errors and the benefits of weak anatomical priors in TargetFusionNet.

\begin{table*}[t]
\centering
\caption{
Comprehensive results across triplet segmentation, detection (bounding box grounding), and recognition (classification).
Mean Average Precision ($\mAP$) is reported for instruments ($\mAP_I$), verbs ($\mAP_V$), targets ($\mAP_T$), instrument–verb pairs ($\mAP_{IV}$), instrument–target pairs ($\mAP_{IT}$), and full triplets ($\mAP_{IVT}$).
\textbf{Bold} indicates the best among compared models.
}
\label{tab:overall_results}
\resizebox{\textwidth}{!}{
\begin{tabular}{llcccccc}
\toprule
\textbf{Evaluation Metric} & \textbf{Method} & $\mAP_I$ & $\mAP_V$ & $\mAP_T$ & $\mAP_{IV}$ & $\mAP_{IT}$ & $\mAP_{IVT}$ \\
\midrule
\multirow{4}{*}{Segmentation $\bm{(\mAP^{seg})}$}
& RDV-Det                 &  0.09 &  0.11 &  0.08 &  0.08 &  0.04 &  0.03 \\
& RDV + Mask2Former       & 48.11 & 32.51 & 16.29 & 14.40 & 11.38 &  8.73 \\
& Mask2Former-Triplet     & 65.24 & 45.61 & 20.75 & 23.03 & 16.47 & 12.23 \\
& \textbf{TargetFusionNet}& \textbf{67.19} & \textbf{46.27} & \textbf{21.55} & \textbf{24.93} & \textbf{17.75} & \textbf{13.47} \\
\midrule
\multirow{4}{*}{Detection $\bm{(\mAP^{det})}$}
& RDV-Det                 &  3.28 &  2.86 &  0.29 &  1.43 &  0.77 &  0.55 \\
& RDV + Mask2Former       & 54.28 & 34.48 & 13.54 & 16.56 &  9.30 &  6.72 \\
& Mask2Former-Triplet     & 64.64 & 45.35 & 20.64 & 22.94 & 16.42 & 12.19 \\
& \textbf{TargetFusionNet}& \textbf{66.52} & \textbf{46.23} & \textbf{23.70} & \textbf{24.82} & \textbf{17.74} & \textbf{13.45} \\
\midrule
\multirow{4}{*}{Recognition $\bm{(\mAP^{rec})}$}
& RDV-Det                 & 80.55 & 61.44 & 40.83 & 35.93 & 34.47 & 27.17 \\
& RDV + Mask2Former       & 80.55 & 61.44 & 40.83 & 35.93 & 34.47 & 27.17 \\
& Mask2Former-Triplet     & 82.70 & 64.64 & 50.22 & 39.79 & 39.93 & 32.28 \\
& \textbf{TargetFusionNet}& \textbf{86.61} & \textbf{71.93} & \textbf{51.19} & \textbf{44.48} & \textbf{43.24} & \textbf{34.23} \\
\bottomrule
\end{tabular}}
\end{table*}

\begin{table}[t]
\centering
\caption{Ablation of target-fusion strategies on segmentation mAP performance. \textbf{Bold} indicates the best among compared models.}
\label{tab:ablation_seg}
\begin{tabular}{lcccccc}
\toprule
\textbf{Method} & $\mAP_{I}^{seg}$ & $\mAP_{V}^{seg}$ & $\mAP_{T}^{seg}$ & $\mAP_{IV}^{seg}$ & $\mAP_{IT}^{seg}$ & $\mAP_{IVT}^{seg}$ \\
\bottomrule
TargetFusionNet  & 67.19 & 46.27 & \textbf{21.55} & 24.93 & \textbf{17.75} & \textbf{13.47}\\
\hline
Early-concat  & \textbf{69.13} & \textbf{50.74 }& 21.52 & \textbf{27.62} & 16.83 & 12.21 \\
Late-concat & 59.90 & 40.65 & 20.21 & 23.17 & 16.07 & 10.99 \\
\hline
\end{tabular}
\end{table}

\mysubsection{Ablations}
To analyse design choices, we compare TargetFusionNet with two simplified fusion variants:  
1) \emph{early fusion}, where weak anatomical logits are concatenated with RGB inputs before the backbone;
and  
2) \emph{late fusion}, where weak anatomical logits are injected after the transformer decoder via a lightweight convolutional encoder.  
Additional analyses in the appendix include comparisons of the standard Mask2Former loss versus multitask loss variants and fusion-depth ablations evaluating how many decoder layers benefit from target fusion.

\mysubsection{Implementation details}
TargetFusionNet was trained in the MMDetection \cite{mmdetection} framework with a ResNet-50 \cite{he2016deep} backbone pretrained for instrument segmentation on data derived from CholecTriplet-Seg, using the AdamW optimiser \cite{loshchilov2017decoupled}. Input images were resized to 1024×1024 with data augmentation comprising random horizontal flipping, cropping, and scaling. Training was performed on a single NVIDIA A100 GPU for 300k iterations, and the best checkpoint was selected based on validation mask mAP.

\section{Results}
\label{sec:results}
\begin{figure}[t]
\centering
\includegraphics[width=\textwidth]{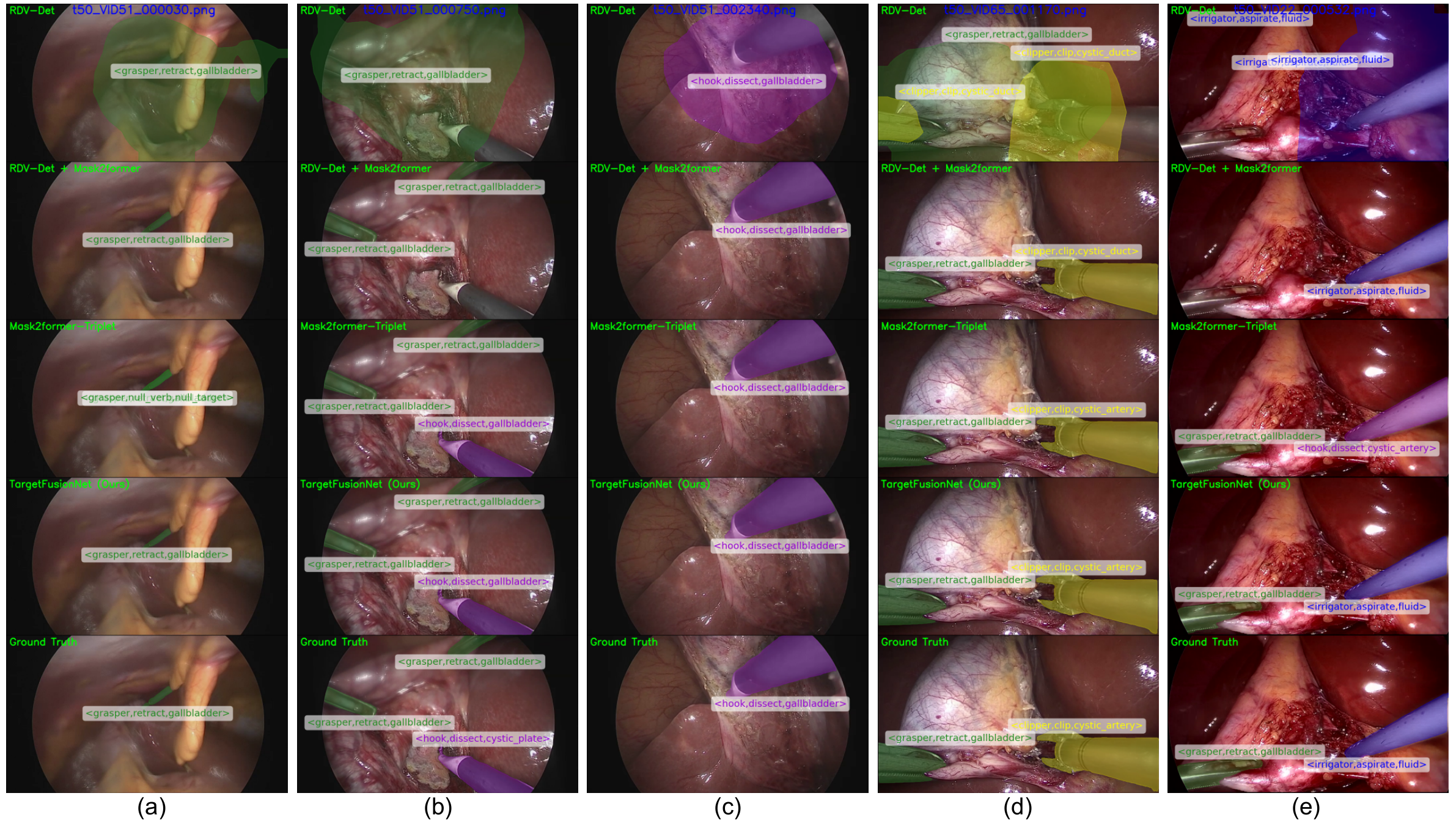}
\caption{Qualitative comparisons across five frames (a–e). Models shown: RDV-Det, RDV-Det + Mask2Former, Mask2Former-Triplet, and TargetFusionNet. Triplet predictions are shown overlaid on the image. Our method achieves more accurate spatial grounding and better triplet consistency in most cases.}
\label{fig:qualitative_results}
\end{figure}

We evaluate RDV-Det, RDV + Mask2Former, Mask2Former-Triplet, and TargetFusionNet on the CholecTriplet-Seg benchmark, reporting recognition (classification), detection (bounding box grounding), and segmentation $\mAP$ as defined in \secref{sec:experiments}.  
Quantitative results are summarised in \tabref{tab:overall_results}, and qualitative examples are presented in \figref{fig:qualitative_results}. Across all evaluation metrics, performance consistently improves with stronger spatial supervision and the introduction of target-aware fusion.  
Segmentation ($\mAP^{seg}$) serves as our primary metric. 
RDV-Det provides negligible instance-level grounding ($\mAP_{IVT}^{seg}=0.03$), while adding strong instrument masks in RDV + Mask2Former raises accuracy to $\mAP_{IVT}^{seg}=8.73$.  
Mask2Former-Triplet achieves $\mAP_{IVT}^{seg}=12.23$, and TargetFusionNet attains the best triplet segmentation accuracy of $\mAP_{IVT}^{seg}=13.47$, with gains across all components, particularly in target prediction ($\mAP_{T}^{seg}=21.55$).

Similar trends are observed in detection and recognition.  
For detection, TargetFusionNet achieves the highest $\mAP_{IVT}^{det}=13.45$, outperforming Mask2Former-Triplet ($\mAP_{IVT}^{det}=12.19$) through more accurate target association.  
For recognition, which ignores spatial grounding, Mask2Former-Triplet improves over RDV baselines ($\mAP_{IVT}^{rec}=32.28$ vs.\ $\mAP_{IVT}^{rec}=27.17$), and TargetFusionNet further increases the triplet score to $\mAP_{IVT}^{rec}=34.23$ (+1.95).

To evaluate robustness, we use $\mAP_{IVT}^{seg}$ and partition the test set into twelve subsets of 500 frames.
A one-sided Wilcoxon signed-rank test shows that TargetFusionNet significantly outperforms Mask2Former-Triplet ($p=0.026$).
This sub-sampling strategy mitigates the small size of the original video-level test set and provides a more stable estimate of instance-level triplet performance.

Looking at the proposed ablations, as shown in \tabref{tab:ablation_seg}, TargetFusionNet outperforms both early and late-fusion variants across most segmentation mAP metrics.

\mysubsection{Qualitative evaluation}
\label{sec:qualitative}
To complement the quantitative analysis, \figref{fig:qualitative_results} compares predictions from RDV-Det \cite{nwoye2022rendezvous}, RDV-Det + Mask2Former, Mask2Former-Triplet, and our TargetFusionNet across diverse scenarios. The examples highlight both strengths and remaining limitations. In case (a) and (c) where there is  a single \triplet{Grasper}{Retract}{Gallbladder} or \triplet{Hook}{Dissect}{Gallbladder} respectively, all models perform reliably, except RDV-Det which has weak localization ability. However, in complex scenes such as (b), (d) and (e) involving multiple active instruments, occlusions, or rare targets, TargetFusionNet achieves more consistent spatial grounding and correct instrument–target linkage. For instance in (d), it successfully distinguishes a \triplet{Clipper}{Clip}{Cystic Duct} from the visually similar cystic artery, and in (e), it recognises the  \triplet{Irrigator}{Aspirate}{Fluid} where baselines misclassify the tool as a hook. These examples illustrate the advantages of target-aware fusion for robust triplet grounding.
A short video comparing all evaluated methods (RDV-Det, RDV-Det + Mask2Former, Mask2Former-Triplet, and TargetFusionNet) with the ground truth is provided in the Supplementary Material.

\section{Discussion and conclusion}
In this work, we introduced the task of triplet segmentation, released the CholecTriplet-Seg benchmark, and proposed TargetFusionNet, which combines strong instance supervision with weak anatomical priors for fine-grained and interpretable surgical scene understanding.
Our experiments demonstrate that models trained with strong instrument supervision achieve the best overall balance across instrument, verb, and target predictions, significantly outperforming weakly supervised approaches such as RDV-Det.
Despite these advances, two key challenges remain.
First, while triplet-optimised models achieve the highest triplet $\mAP_{IVT}^{seg}$, their instrument masks ($\mAP_{I}^{seg}$) are slightly less precise than those from single-task segmentation models \cite{alabi2025cholecinstanceseg}.
This trade-off highlights the need for hybrid or multi-stage architectures that preserve mask quality while enabling effective triplet reasoning.
Second, target segmentation still remains a bottleneck, as low target $\mAP$ directly constrains overall triplet accuracy.
Improving anatomical modelling beyond weak priors could further enhance spatial grounding and downstream reasoning.
Future extensions could also incorporate temporal reasoning to exploit action continuity across frames and integrate graph-based or relation-aware modules to capture richer inter-instrument and tissue relationships.

\section*{Declarations}
\noindent\textbf{Open access} Code and dataset are publicly available from Github,
\url{https://github.com/labdeeman7/target_fusion_net} 
and Synapse
\url{https://doi.org/10.7303/syn70783776.1} respectively.

\noindent\textbf{Declaration of Interests} 
TV is a co-founder and shareholder of Hypervision Surgical Ltd, London, UK.
The authors declare that they have no other conflict of interest.

\noindent\textbf{Ethics approval} For this type of study, formal consent is not required.

\noindent\textbf{Funding Sources} This work was supported by core funding from the Wellcome/EPSRC [WT203148/Z/16/Z; NS/A000049/1] and, Tongji Fundamental Research Funds for the Central Universities. OA is supported by the EPSRC CDT
[EP/S022104/1].

\bibliography{sn-bibliography}

\end{document}